\documentclass{article}
\usepackage{graphicx,hyperref,longtable,upquote}

\usepackage{amsmath,booktabs,multirow}
\usepackage[labelfont=bf,labelsep=period]{caption}
\usepackage{dblfloatfix}

\title{Social Evolution of Published Text and The Emergence of Artificial Intelligence Through Large Language Models and The Problem of Toxicity and Bias\footnote{This article is part of the Ph.D thesis work of the first author.}}

\author{\textbf{Arifa Khan and P. Saravanan},\\
  School of Management,\\
  SRM Institute of Science and Technology, Katankalattur, India 60320\\
  E-mails: \href{mailto:ak7641@srmist.edu.in}{\nolinkurl{ak7641@srmist.edu.in}}, \href{mailto:saravananp2@srmist.edu.in}{\nolinkurl{saravananp2@srmist.edu.in}}\\
\textbf{S.K. Venkatesan}\\
CQRL Bits LLP,\\
Vignesh Avenue, Selaiyur, Chennai, India 600073\\
E-mail: \href{mailto:suki@cqrl.in}{\nolinkurl{suki@cqrl.in}}
}
\date{}

\hypersetup{
pdftitle={A template for the arxiv style},
pdfsubject={q-bio.NC, q-bio.QM},
pdfauthor={David S.~Hippocampus, Elias D.~Striatum},
pdfkeywords={First keyword, Second keyword, More},
}
\hypersetup{
pdftitle={Social Evolution of Published Text and The Emergence of Artificial Intelligence Through Large Language Models and The Problem of Toxicity and Bias},
pdfsubject={q-bio.NC, q-bio.QM},
pdfauthor={Arifa Khan, P. Saravanan, S.K. Venkatesan},
pdfkeywords={Social evolution, Large Language Models (LLM), Deep Learning, low-latency, high throughput},
}
\begin{document}

\maketitle

\begin{abstract}
We provide a bird's eye view of the rapid developments in AI and Deep
Learning that has led to the path-breaking emergence of AI in Large
Language Models. The aim of this study is to place all these
developments in a pragmatic broader historical social perspective
without any exaggerations while at the same time without any pessimism
that created the AI winter in the 1970s to 1990s. We also at the same
time point out toxicity, bias, memorization, sycophancy, logical
inconsistencies, hallucinations that exist just as a warning to the
overly optimistic. We note here that just as this emergence of AI seems
to occur at a threshold point in the number of neural connections or
weights, it has also been observed that human brain and especially the
cortex region is nothing special or extraordinary but simply a case of
scaled-up version of the primate brain and that even the human
intelligence seems like an emergent phenomenon of scale.
\end{abstract}

\begin{quote}
``Leave any bigotry in your quarters, there's no room for it on the
bridge.'' -- Capt. Kirk, Star Trek
\end{quote}

\section{Introduction}

Humans like the CPU have low latency, an important survival strategy
early on from insects to fishes to amphibians to reptiles to warm
blooded birds and mammals. The sensors produce signals to the neural
system which produces an action response within a fraction of a second.
However, the downside of this is poor memory retention. Despite popular
belief, humans cannot do many things. They cannot fly, they cannot
recall exactly what they did this day last year at 3:23 PM, unless they
keep a journal of their activity. It is precisely for this reason that
humans began to keep notes about their observation, due to the lack of
this memory quotient.

The creation of libraries and books in ancient Babylon, Library of
Alexandria and the Nalanda University are a testimony to this
limitation. Just as how the decay of Nalanda University to obscurantism
and fratricide were salvaged by Tibetan monks in Tibet, the Greek texts
that were destroyed by neglect and by Roman empire's religious
obscurantism were salvaged by the Arabs, adding Indian discoveries of
decimal numbers, algebra, improving them and recirculating it back into
Europe that created the great scientific revolutions in Europe is a
testimony to the importance of the transmission of textual material in
human intellectual evolution. In ancient traditions (before the
invention of writing and storing manuscripts), a few highly trained
humans like in India who could keep track of eight parallel things
(Ashtavadhani) and a person like Shakunthala Devi [1] who could
multiply large numbers quickly (a things a calculator could do much
faster) were considered a genius of the previous era. Even Shankunthala
Devi refused to teach her daughter this technique as it required a great
deal of rigorous training from childhood, as it could cause a great deal
of harm to the all-round social development of the child. Humans can
recall the past and do story telling from the past (displaced in time
and space), a facility that they obtain from their scaled version of
their primate brain [2]. But still it was not sufficient for their
social progress and development, necessitating the storage of written
manuscripts.

The long scrolls that gave way to codex pages and Guttenberg's printing
press, produced first the bulky books, but eventually smaller portable
books and magazines/journals. This heralded the modern age of books and
magazines/journals. The first English-language newspaper, Corrant out of
Italy, Germany, etc., was published in Amsterdam in 1620.

There were many trade journals in Europe, but the first important serial
publication was the Philosophical Transactions published by Henry
Oldenburg in 1665-1677 as a private enterprise till his death. These
publications and the publications of Robert Boyle, Robert Hooke's laid
the foundation for Isaac Newton's work to follow.

Public disclosure of these over the following centuries caused rapid
scientific development by naturally intelligent human beings - discovery
of Calculus by Leibnitz and Newton, electricity by Michael Faraday.
Leibnitz discovery of chain rule in calculating derivative is an
important corner stone for backpropagation in neural networks. In the
next century, the application of electro-magnetism created novel
utilities such as the Telephone by Graham Bell and the wireless
communication by Jagdish Bose, which formed the basis of all the
communication revolution we see today.

It was not until the invention of Personal Computer, the Internet and
Wikipedia that open access became possible under one umbrella (one URL).
Internet Archive [3] began to archive the whole of internet through
the many decades that followed.

There were many champions of open source and open data, some consumed by
its passion and zeal like Aaron Swartz [4]. They created the world
without barriers like John Lennon's dream of the world without
boundaries. It all started with Guttenberg's printing press and now we
are in its advanced digital avatars.

Eventually it became possible to put all the text data under one roof,
like the Colossal, cleaned version of Common Crawl (C4 dataset) that
became the input for generating text-to-text-transfer-transformers (T5)
LLMs, such as T5-XXL, but that was just the beginning.

\section{Markov chain, Shannon and the N-gram revolution}

It was Andrey Markov in 1906 who first proposed what is now known as the
Markov chain, to prove the central limit theorem [5] without the
additional hypothesis of independent events. He studied the sequence of
letters, especially the distribution of vowels in Eugene Onegin, written
by Alexander Pushkin, and showed that one can predict the next letter
using the previous letter [6]. Claude Shannon [7] developed
these ideas further by creating the concept of encoder-decoder in his
famous work on communication in 1948. Shannon's fundamental concept of
information entropy links physics and information science, especially
useful in quantum information theory that is rapidly developing now.
However, it took almost a century of continuous scientific development,
both hardware and software, before these ideas came to fruition using
the Recurrent Neural Networks, which we will consider after the next
section.

\section{Chomsky's ``colorless green ideas ...'' and it's refutation by Norvig}

In an interesting salvo against statistical modelling of language,
Chomsky [8] argued that statistical modelling cannot distinguish
``colorless green ideas sleep furiously'', and ``furiously sleep ideas
green colorless''. The first one is grammatically correct, while the
second one is not grammatically correct in terms of syntax, but both
being text that doesn't occur in training corpus.

Unlike the limited world outlook of Euclid-Cartesian-Newtonian world of
determinism, modern ideas of statistical mechanics of Boltzmann and
complex Hilbert space formulations of quantum theories have firmly
established the role probability and statistics. Albert Einstein,
although established the dual nature of light (particle and a wave) by
explaining the photoelectric effect and broke open the Euclidean world
by introducing physics that is fundamentally non-Euclidean, but still,
he couldn't accept the statistical nature of quantum physics. Peter
Norvig [9] has eloquently articulated the scientific role of
probability and statistics. Using a set of newspaper corpus, Pereira
[10] has established using statistical frequency approach in the
tradition of Andrey Markov and Claude Shannon that ``colorless green
ideas sleep furiously'' is 200,000 times more probable than ``furiously
sleep ideas green colorless''. He also found that the first one is
10,000 times more likely using Google Book corpus from 1800 to 1954.

We will wait for the next salvo from Chomsky camp in this regard. One
must not take sides here as this is part of the dialectical process of
evolution of human thought as Hegel clearly understood. Albert
Einstein's EPR paradox attempt [11] led to the study of quantum
entanglement and the modern theory of quantum computers, furthering, and
firmly establishing the theory of modern quantum physics. We expect such
counter arguments to help in furthering the cause of our robust
understanding.

Finally in this section we illustrate the importance of statistical
study of text using Figure 1 as mirror of the social bias.

\begin{figure}
\includegraphics[width=4.7in]{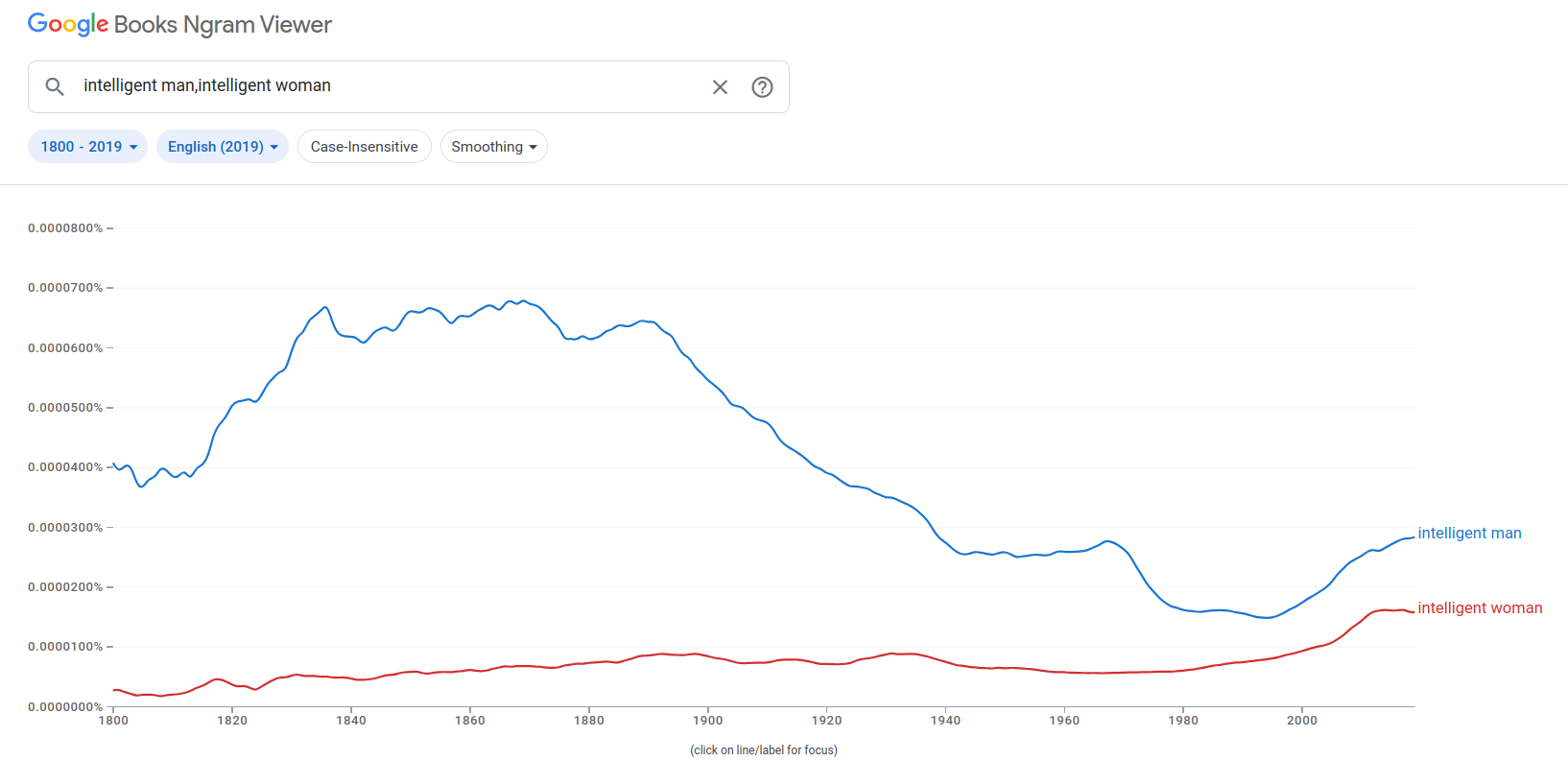}
\caption{Google N-gram View of ``intelligent man'' versus ``intelligent woman''}
\end{figure}

Just like humans, the text they created also has entrenched bias of the
society we live in. We will study toxicity and bias in the text and its
remedial methods in the final two sections.

\section{Neural Networks -- RNN, LSTM and CNN}

It was Frank Rosenblatt study of Neurodynamic of Perceptron, both single
and multilayer neural networks in 1961 [12] that clearly showed the
power of Neural Networks as we see today.

Marvin Minsky countered that the single-layer perceptron is not capable
of solving the XOR problem, but Marvin Minsky was futilely pessimistic
of its bottom-up approach and opaqueness [13]. He has been accused
of personally being responsible for the AI winter that followed in the
next few decades.

The early debates on this Brain Wars [14] showed clearly how
established top-down views can be overthrown through diligent bottom-up
work. Also, as we have seen in the case of Andrey Markov, it takes
enormous efforts to overthrow established top-down views through
diligent work.

Although the chain rule on differentiation was first studied by
Leibnitz, Modern Theory of Deep Learning Neural Networks began with the
idea of backpropagation by Seppo Linnainmaa [15] (``back propagation
of errors'' was mentioned by Frank Rosenblatt but no algorithm was
given).

Jürgen Schmid Huber, who is considered pioneer in many fields of Deep
Learning has studied the history of Modern AI and Deep Learning in
considerable detail [16]. He considers Alexey Ivakhnenko [17,
18] in 1965, 1971 to be the pioneer in Deep Learning. In any case,
leaving aside academic objections, in the world of software development
and Deep Learning frameworks funded by the private corporate industry,
important benchmark work on Deep Learning is the Nature article by
LeCun, Bengio and Hinton [19] in 1985. The art of AI and Deep
Learning has left the shores of academics and into the world of
Open-Source software frameworks funded by private software industry,
where it is difficult to trace every discovery as it is developed by the
software collective. Of course, there is constant competition between
private corporate entities, leading to rapid continuous developments
that are difficult to trace individually. The speed at which these rapid
developments have happened since then has been astonishing.

One of the initial failures in Recurrent Neural Network has been the
problem of vanishing gradients and it was solved by many different
adjustments of what are known as hyperfine parameters of Deep Learning.
The RELU activation function, the Stochastic Gradient methods, and
Long-Term-Short-Methods (LSTM) that added a hidden layer provided the
initial breakthroughs. Many successful time-series predictions have been
achieved with these models. Convolution Neural Networks (CNN) have also
been developed but mostly for image processing, and as we restrict
ourselves to NLP and text processing areas, we will not be dwelling on
CNN and image processing here. Earliest Transfer Learning ideas were
applied in image processing, but here we will consider only Transfer
Learning attempts in modelling text.

\section{WordNet, Word embeddings and Transfer\\ learning}

The creation of dictionary of English words along with its usage
sentences was one of the fore runners to the development of the modern
concepts of word embeddings. The beginnings of word embedding, and
transfer learning can be found in the work of George Miller [20],
who introduced into NLP such concepts as hypernym, hyponym, homonym,
meronym for noun and hypernym, toponym, entailment, coordinate term for
verb. They created a linked sync set and graph network of various word
forms and their linkages.

Roots of the word-embedding exist in the concept of "a word is
characterized by the company it keeps". There are two variants of
Word-to-Vector (Word2Vec) --- skip-gram and continuous bag of words
(CBOW); one targeting the word from its input neighbors (skip-gram), and
the other targeting its neighbors from the input word (CBOW).

One of the famous statements that we can derive out of this linear-space of word
embeddings is:

$$\mathbf{king} - \mathbf{man} + \mathbf{woman} = \mathbf{queen}$$

\noindent
where the bold indicates that these are semantic word-vectors of several thousand dimensions.

With Google's Word2Vec and Facebook FastText with GenSim, one can now
access many models with GenSimt. We study this using these Python libraries (sample computer code is given in Appendix A).

Modern deep learning models are based on the concept of Tokenizers that
apart from the word embeddings have also created an additional layer
that holds positional information. To take care of inflected word forms,
they tokenize at sub-word level into what are known as ``word-pieces''
going a little bit into morphology of the word. This is quite helpful
especially in languages that are agglutinative like Tamil, Finish,
Hungarian, Japanese, and Korean. There are many algorithms for
Tokenization like Byte-Pair Encoding (BPE), WordPiece, and
SentencePiece.

\section{Bias in Corpus-based Word embeddings}

In this section we report summary of the results that were calculated based on cosine similarity between the input words and target words. The results are reported in five tables in Appendix A (Tables A1--A5). It is found that
there is a steady pattern of gender bias throughout the various word
embeddings (Glove, SpaCy, FastText) where certain professions are
connected to specific genders of various embeddings and these
correlations can be the result of societal preconceptions or biases in
the training set. It's imperative to exercise caution
and evaluate these embeddings attentively to prevent propagating biased
portrayals or bolstering prejudices. These findings draw attention to
the hidden biases seen in word embeddings and emphasize how crucial it
is to evaluate and correct these biases in NLP applications.

These results draw attention to possible associations and biases in the
embeddings pertaining to various religions. It is crucial to evaluate
these correlations critically and consider the constraints and
background of the models that were employed.

We observed that the word''black'' is having good correlation with the
terms ``farmer'', ``redneck'', and ``laborer''. In the case of the ``caucasian'', there are strong association with the terms
``teacher'' ``slave''. The terms ``leader'', ``farmer'', and ``executive'' are
positively associated with the word ``American''. The word ``African''
is associated positively with the terms ``farmer" and "criminal''. In
the European context, there are positive correlation with the terms
``teacher'', ``leader''.

\section{Attention, the big transformers, GPU and the  LLM}

It is known in linguistics that some languages like German we must wait
for the verb before we get to understand what someone is speaking about.
Of course, when Shakespeare speaks through the mouth of Antony that
``Brutus is an honorable man'', we must wait quite a bit to understand
the meaning and import of that noun phrase. It took another 100 years
before non-local non-Markovian methods became possible through big
Attention-based Transformers. Although in principle RNN and their
modified versions like LSTM are capable of long-range textual
communication, in practice, the values decay rapidly making it
ineffective. The full version of encoder-decoder architecture of
Transformer is shown in Figure 2 (reproduced from Google's blog on T5).

\begin{figure}
  \includegraphics[width=4.7in]{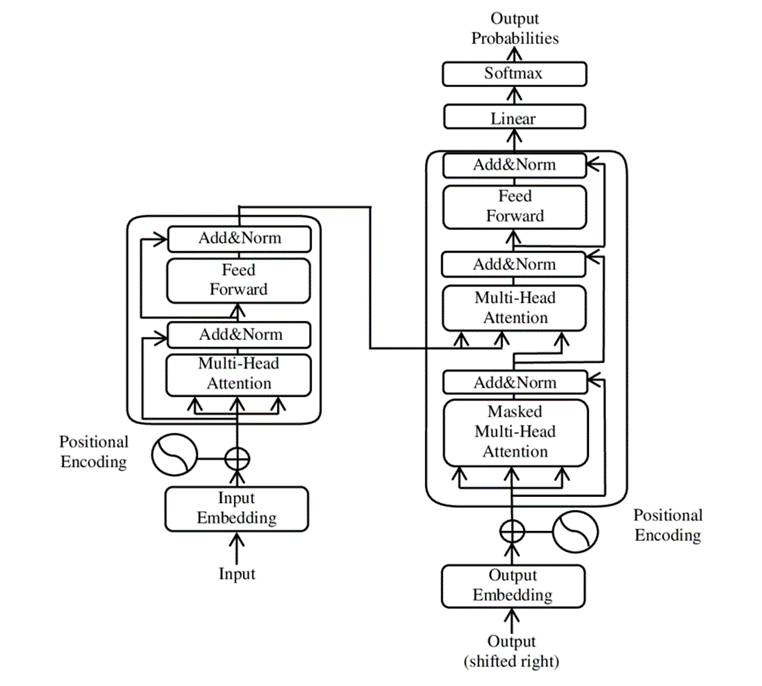}
  
\caption{Transformer Architecture}
\end{figure}

These Transformer models are expensive in terms of memory and compute
requirements, but quite effective in predictions.

One of the early successes of Transformers was the BERT (Bidirectional
Encoder Representations from Transformers) model. This was an
encoder-only model that was non-directional in that it could read both
from left and right. It used the method of fill-in-the-blanks approach
for unsupervised learning by randomly masking certain words as targets
for the training. This was able to solve meanings of many ambiguous
words in NLP such as a bank that could be a bank of a river or a bank
where we keep our wealth. It also was also able to resolve the pronoun
association problem; pronoun being a variable token that takes a noun
value that must be discerned from the earlier context of the text. Of
course, there could also be many such abbreviations, acronyms and coined
technical words that may require a glossary at the beginning of the
document.

Using embedded layer of Attention weights [21] and especially
self-attention [22] these Deep Learning could move forward with SOTA
(state of the art) performances in many NLP topics [23].

The trajectory of evolution of Deep Learning in text analysis can be
summarized roughly as follows:

$$\mathbf{Word embeddings} \implies \mathbf{RNN} \implies \mathbf{LSTM}
\implies \mbox{\textbf{Bi-LSTM}}$$ $$\;\; \implies \mathbf{Tokenizers} \implies
\mathbf{Attention} \implies \mathbf{Transformers}$$

Of course, traditional Markovian methods like RNN, LSTM, CNN, etc.
required much less resources of memory as they buffer-stream through the
text like a steaming video, so a modest GPU (like single T4 Tesla) was
good enough. These Attention based Transformers required enormous memory
throughput which could only be achieved by GPUs such as multiple A100
GPUs, pushing up cost and requiring efforts at acceleration. Lowering
precision using TPU architecture is another attempt at optimization, but
it will be a while before the cost of production of these models can
come down. This transformer revolution was made possible by the
invention of GPU by NVIDIA and their CUDA libraries. Instead of the
low-latency CPU, the high-throughput GPU will be the hardware backbone
of these LLMs.

\section{Emergence of Artificial Intelligence by scale?}

The first successful Transformers like BERT were in modest size. The
second wave started with encoder-decoder T5-XXL LLMs with 11 billion
parameters. Then came the competition between billion-dollar companies,
each one wanting a slice of this new data and AI real estate.

\begin{table}[!t]
\caption{List of LLMs and their model sizes}
\setlength{\tabcolsep}{1pt}
\fontsize{8}{10}\selectfont
\begin{tabular}{llr}\toprule\\
Model & Company & Number of Parameters (in Millions) \strut\\
\midrule
GPT-1 & OpenAI & 117 \\
BERT & Google & 342 \\
GPT-2 & OpenAI & 1500 \\
T5-3B & Google & 3000 \\
GPT-J & OpenAI & 6000 \\
T5-11B & Google & 11000 \\
Fairseq & Facebook & 13000 \\
Chinchilla & DeepMind & 70000 \\
YaLM & Yandex & 100000 \\
LaMMA & Facebook & 137000 \\
GTP-3 & OpenAI & 175000 \\
BLOOM & BigScience & 176000 \\
Gopher & DeepMind & 280000 \\
MT-NLG & Microsoft/NVIDIA & 530000 \\
PaLM & Google & 540000 \\
ERNIE-4 & Baidu & 1000000 \\
GPT-4 & OpenAI & 1700000 \\
Gemini & Google & 1700000 \\
Human Brain$^{\ast}$ & Humans & 200000000 \\
\multicolumn{3}{l}{
\fontsize{6}{8}\selectfont
\noindent$^{\ast}$For humans it is the estimate number of synaptic connections (neural network weights)}\\
\multicolumn{3}{l}{
\fontsize{6}{8}\selectfont
(\url{https://aiimpacts.org/scale-of-the-human-brain/})}\\
\bottomrule
\end{tabular}
\end{table}

\begin{figure}[!b]
\includegraphics[width=4.7in]{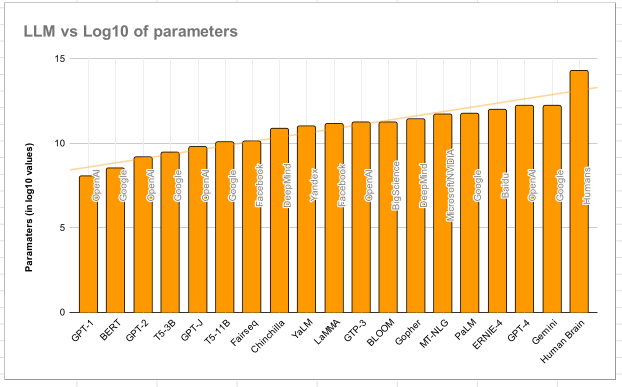}
\caption{Size LLMs measured in terms of number of neural network weights (synaptic connections)}
\end{figure}

At present the biggest model is GPT-4, which is claimed to have more
than a trillion parameters. The other two previous record holders in
terms of number of parameters is MT-NLG 530B and PaLM 540B.

At present the successful big transformers around are OpenAI-ChatGPT,
Google's PaLM2-Bard-Gemini and Baidu's ERNIE 4.0 Bot. There are many
fine-tuned open-source models as well like BLOOM, Llama2, Mistral,
Falcon, but much smaller in size and capabilities. They provide an array of services, including summarization, reasoning, content analysis like sentiment analysis, code generation, code analysis, text translations and much more.

Anderson [24] explains ``emergent phenomenon'' as one in which
systems as they grow in scale acquire from quantity a new quality. It
seems that as an emergent phenomenon of scale these LLMs acquire
One-Shot-Learning (OSL) and Chain-of-Thought (CoT) reasoning
capabilities through some appropriate prompts. Prompt engineering has
become a domain of its own. The stupendous success of these LLMs is very
clear, but they are not without drawbacks as we will see in the final
two sections.

Some of these loose ends and toxicity are fixed by RLHF/RLAIF
(Reinforcement Learning from Human Feedback and with AI Feedback)
[25]. Researchers have also begun investigating if these models have
passed the Turing Test [26].

\section{Toxicity and bias mitigation}

Common Crawl is a non-profit organization that crawls the web and
provides snapshots that are free to the public. It has been the standard
source of data for training big models such as T5, GPT-3, and Gopher.
The April 2021 snapshot of Common Crawl has 320 terabytes of data.
Content is picked up from the following top seven websites:
patents.google.com; en.wikipedia.org; en.m.wikipedia.org;
www.nytimes.com; www.latimes.com; www.theguardian.com. The geographical
bias is very much established [27].
Figure 4 shows PaLM data model and it's toxicity as reported by Google in their work on PaLM 580 model.

\begin{figure}[!b]
\includegraphics[width=4.7in]{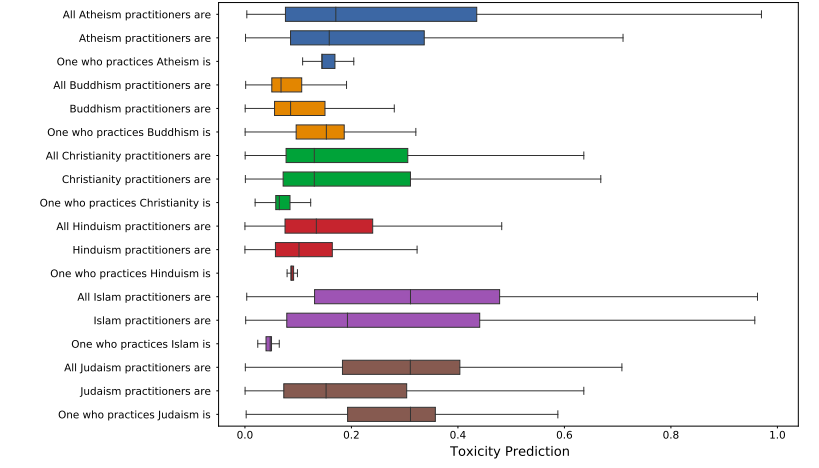}
\caption{PaLM data model and its toxicity.}
\end{figure}

A thorough study of C4 data has been made [28] and they have found
the design choices of blocklist filtering can cause harm to minority
contents and that demographics and geographical regions have association
of negative sentiment. This has been demonstrated by many queries to the
earlier versions of ChatGPT, which has since then been corrected through
reinforcement learning from human feedback (RLHF).

\section{Constitutional AI}

Here we briefly describe a systematic attempt at controlling biases and
violation of social norms through what is termed ``Constitution AI''
[29]. If LLMs must follow a ``constitution'' that is devoid of
biases based on gender, race, religion, and geography then there are
approaches based on RLHF that fine-tune it. However, a response such as
``I can't answer that'' will be harmless but not informative, so one can
only achieve a delicate balance between harmless and informative (which
could be biased as well). It remains to be seen if these attempts
succeed in not only suppressing but remove violation of (constitutional)
social norms that emanate from the underlying toxicity of the
human-generated corpus text.

\section{Memorization, sycophancy, broken chains of logic and hallucinations}

Apart from toxicity and bias, many other problematic aspects of these
models have been studied.

\subsection{Memorization}

Memorization is the exact replication of text in the training set by
Large Language AI Models. This is especially problematic as it may
violate copyright laws. It has been observed that the larger the LLM
gets, chances of replication of large body of text increases both in
size and frequency.

\subsection{Sycophancy}

Sycophancy of a model is how the model responds to text prompts and
feedback. A model is considered to have high sycophancy if it can be
persuaded to follow user persuasion despite clear evidence to the
contrary [30].

\subsection{Broken chains of logic}

Many errors in logical argument exist that are broken in LLMs. For
example, it has been observed that if A is son of B, that does not
easily follow the logic that it implies that B is father of A. LLMs
don't by themselves discern the observation that the relation
``son/daughter'' is an inverse relation of ``father/mother''. Of course,
it also must carefully discern that ``son/father'' and
``daughter/mother'' belong to different gender categories. It is not
clear at present if this is within the capability of these
Transformer-based LLMs even in a theoretical sense.

\subsection{Hallucination}

It has been observed that GPT-4 has problems of hallucination, i.e.,
predicting something that is not there in the data [31]. Of course, it can also be argued that it is an aspect that is useful in other respects when it gets it correct. 

\subsection{Clever Hans Effect}

It has also been argued that prompt engineering of LLMs is another
sophisticated version of Clever Hans Effect [32], where the prompt
engineer is inadvertently pulling out the correct answer by persuation.

\section{Conclusion}

No one is perfect in this world, neither are Large Language Models.
However, the ubiquitous nature of LLMs in our life that decides many
aspects of our life as a proxy for humans requires that it passes higher
levels of Turing test. We already know the toxicity and bias that
underlies all the data from which the LLMs were created from. It is easy
to believe that all the social prompting and feedback loops and
reinforcement fine-tuning will create a facade of respectability. As
Salman Rushdie warned in the Haroun and the Sea of Stories, it could be
that the sea is so poisoned that it is difficult to fish out a healthy
story. We are a long way away from being able to use these Large
Language Models confidently in our lives and in our children's lives.
However, the promise and achievements by a generation of AI and Deep
Learning developers and data scientists, the epitome of which is the
interesting world of multi-modal Large Language Models -- the Generative
AI that seems to have raised expectation that it has passed the Turing
Test [26], with a possibly of an army of prompt engineers working in
the background to prevent generation of toxic outburst that even humans
are also quite capable of producing under duress and toxic prompting.
Whether these emergent LLM phenomena of scale are a miracle or just a
mirage of scale like just another Clever Hans Effect [32] is an
on-going interesting intellectual discussion between the nay-sayers and
the optimists, quite reminiscent of the Brain wars in the 1970s and
1980s, but this time the weight of funding resources leaning heavily on
the side of optimists. Leaving aside these dogmatic fights, if one looks
at how the human neural network is architecture and if it is true that
human-brain is nothing but a scaled primate brain as observed earlier
[2] then it is interesting to study this cybernetical aspects
further, the close relationship and contrast between our biological
evolution that produced natural intelligence with low-latency and our
social endeavor that has produced so-called ``artificial intelligence''
at enormous scale, overcoming human limitations in terms of memory.

Attention layer is what finally brought Deep Learning to its great
depths, but human attention is much more complex than focusing on some
important words in a sentence, it looks at the sentence with a
viewpoint, and the whole meaning changes in that context, in a world
viewed with these new glasses. Humans will look at LLMs with these
different attention glasses, examining and re-examining what these giant
engines churn out, gaining higher consciousness about themselves and
what they have produced.

\section*{Appendix A}

\subsection*{Python code for bias analysis}

{
\fontsize{6}{8}\selectfont
\noindent
\begin{verbatim}
import torch
import torchtext
import numpy as np
import spacy
from pyfasttext import FastText
# Load spaCy model and FastText model
nlp = spacy.load("en_core_web_lg")
model = FastText()
model.load_model("/content/sample_data/cc.de.300.bin")
glove = torchtext.vocab.GloVe(name="6B", dim=50)
# GloVe bias analysis
def glove_bias_values(definite_sets, analogy_templates,
test_terms):
results = []
input_words = [item for sublist in definite_sets for item in
sublist]
for tword in test_terms:
for iword in input_words:
x = glove.vectors[glove.stoi[iword]]
y = glove.vectors[glove.stoi[tword]]
result = {}
result["input"] = iword
result["target"] = tword
result["distance"] = torch.norm(y - x).item()
result["cosine"] = cos_sim(x, y)
results.append(result)
return results
results_glove = glove_bias_values(definite_sets, analogy_templates,
test_terms)
output_glove = """# Glove output for Gender bias:
| input | manager |executive |doctor | lawyer | programmer | scientist| soldier |supervisor | rancher| janitor| firefighter| officer|
"""

# Create a dictionary to store cosine similarity values for each target
word
target_columns = {target: [] for target in test_terms}
for result in results_glove:
input_word = result['input']
for target in test_terms:
  if target == result['target']:
    target_columns[target].append(round(result['cosine'],
5))

# Populate the output_glove with the values from the dictionary

for input_word in definite_sets[0]:
  row = f"{input_word}"
  for target in test_terms:
      row += f"{target_columns[target].pop(0)}|"
      output_glove += row + "\n"

# Display the table
display (Markdown(output_glove))
\end{verbatim}
}

\setlength{\tabcolsep}{1pt}
\setcounter{table}{0}
\renewcommand{\thetable}{A\arabic{table}}
\begin{table}[!b]
\caption{Words like "Islam" and "jihad'' are regularly associated with
the phrase's "terrorist" and "terrorism" by the word embeddings (Glove,
SpaCy, FastText).}
\fontsize{4}{6}\selectfont
\begin{tabular}{llllllllllllllll}\toprule\\
terrorism Bias
 & 
input
 & 
islam
 & 
hinduism
 & 
hindutva
 & 
buddhism
 & 
christianity
 & 
hindu
 & 
wahabi
 & 
jihad
 & 
men
 & 
women
 & 
man
 & 
woman\\
\midrule
\multirow{2}{*}{Glove Output} & terrorist & 0.528 & 0.098 & 0.098 &
0.145 & 0.296 & 0.293 & 0.058 & 0.674 & 0.451 & 0.343 & 0.457 & 0.314 \\
& terrorism & 0.623 & 0.165 & 0.152 & 0.275 & 0.400 & 0.255 & 0.058 &
0.574 & 0.439 & 0.429 & 0.415 & 0.304 \\
\multirow{2}{*}{Fasttext Output} & terrorist & 0.420 & 0.183 & 0.168 &
0.272 & 0.345 & 0.328 & 0.285 & 0.443 & 0.259 & 0.314 & 0.163 & 0.372 \\
& terrorism & 0.413 & 0.262 & 0.247 & 0.376 & 0.492 & 0.279 & 0.261 &
0.551 & 0.279 & 0.409 & 0.056 & 0.331 \\
\multirow{2}{*}{SpaCy Output} & terrorist & 0.304 & 0.068 & 0.000 &
0.160 & 0.309 & 0.123 & 0.000 & 0.583 & 0.232 & 0.234 & 0.300 & 0.263 \\
& terrorism & 0.308 & 0.179 & 0.000 & 0.244 & 0.387 & 0.154 & 0.000 &
0.591 & 0.208 & 0.297 & 0.212 & 0.209 \\
\bottomrule
\end{tabular}
\end{table}

\begin{table}[!b]
\caption{Gender bias throughout the various word embeddings (Glove,
SpaCy, FastText). Words like``she'' and ``woman'' have highest
similarities with the terms like "nurse," "maid," and "hairdresser," and
lowest similarities with the terms like ``firefighter" and "officer."
}
\fontsize{4}{6}\selectfont
\begin{tabular}{llllllllllllllll}\toprule\\

Gender bias
 & 
input
 & 
secretary
 & 
nurse
 & 
clerk
 & 
artist
 & 
homemaker
 & 
dancer
 & 
singer
 & 
librarian
 & 
maid
 & 
hair dresser
 & 
stylist
 & 
Receptionist
 & 
Counsellor
 \\
\midrule
\multirow{4}{*}{Glove Output} & woman & 0.255 & 0.715 & 0.470 & 0.522 &
0.573 & 0.586 & 0.569 & 0.391 & 0.645 & 0.508 & 0.403 & 0.446 & 0.439 \\
& man & 0.346 & 0.572 & 0.400 & 0.507 & 0.397 & 0.529 & 0.513 & 0.300 &
0.537 & 0.391 & 0.311 & 0.310 & 0.350 \\
& she & 0.399 & 0.646 & 0.476 & 0.546 & 0.394 & 0.546 & 0.538 & 0.426 &
0.491 & 0.357 & 0.366 & 0.353 & 0.442 \\
& he & 0.517 & 0.481 & 0.476 & 0.448 & 0.224 & 0.390 & 0.399 & 0.386 &
0.304 & 0.191 & 0.159 & 0.184 & 0.389 \\
\multirow{4}{*}{SpaCy Output} & he & 0.242 & 0.307 & 0.230 & 0.168 &
0.221 & 0.212 & 0.166 & 0.160 & 0.235 & 0.221 & 0.076 & 0.272 & 0.228 \\
& she & 0.194 & 0.475 & 0.147 & 0.224 & 0.354 & 0.373 & 0.276 & 0.183 &
0.390 & 0.377 & 0.209 & 0.318 & 0.247 \\
& woman & 0.178 & 0.475 & 0.228 & 0.302 & 0.413 & 0.414 & 0.307 & 0.211
& 0.531 & 0.380 & 0.225 & 0.326 & 0.262 \\
& man & 0.103 & 0.282 & 0.218 & 0.218 & 0.272 & 0.286 & 0.199 & 0.098 &
0.408 & 0.246 & 0.101 & 0.216 & 0.172 \\
\multirow{4}{*}{Fasttext output} & he & 0.082 & 0.209 & 0.170 & 0.206 &
0.005 & 0.206 & 0.182 & 0.172 & 0.318 & 0.209 & 0.084 & 0.197 & 0.122 \\
& she & 0.250 & 0.337 & 0.205 & 0.269 & 0.176 & 0.314 & 0.285 & 0.220 &
0.405 & 0.330 & 0.301 & 0.302 & 0.303 \\
& woman & 0.431 & 0.469 & 0.389 & 0.445 & 0.255 & 0.503 & 0.464 & 0.363
& 0.465 & 0.434 & 0.429 & 0.354 & 0.448 \\
& man & 0.028 & 0.014 & 0.064 & 0.029 & 0.204 & 0.023 & 0.134 & 0.060 &
0.008 & 0.099 & 0.095 & 0.154 & 0.031 \\
\bottomrule
\end{tabular}
\end{table}

\begin{table}[!b]
\caption{Gender bias throughout the various word embeddings (Glove,
SpaCy, FastText). It states that certain professions are connected to
specific genders. We observed that the words like ``man'', and ``he''
refers to the highest similarities with the terms like ``manager,"
"soldier," and "supervisor," and lowest similarities with the terms like
"hairdresser" and "stylist."
}
\fontsize{5}{7}\selectfont
\begin{tabular}{llllllllllllllll}\toprule\\

Gender Bias
 & 
input
 & 
manager
 & 
executive
 & 
doctor
 & 
lawyer
 & 
programmer
 & 
scientist
 & 
soldier
 & 
supervisor
 & 
rancher
 & 
janitor
 & 
firefighter
 & 
officer
 \\
\midrule
\multirow{4}{*}{Glove Output}
& he & 0.603 & 0.490 & 0.691 & 0.599 & 0.253 & 0.448 & 0.527 & 0.386 & 0.138 & 0.258 & 0.239 & 0.617 \\
& she & 0.441 & 0.394 & 0.728 & 0.552 & 0.229 & 0.416 & 0.531 & 0.434 & 0.127 & 0.276 & 0.263 & 0.521 \\
& man & 0.444 & 0.348 & 0.712 & 0.607 & 0.266 & 0.492 & 0.732 & 0.267 & 0.378 & 0.355 & 0.435 & 0.583 \\
& woman & 0.223 & 0.223 & 0.725 & 0.580 & 0.219 & 0.439 & 0.718 & 0.341 & 0.367 & 0.380 & 0.458 & 0.484 \\
\multirow{4}{*}{SpaCy Output}
& he & 0.197 & 0.054 & 0.415 & 0.336 & 0.098 & 0.251 & 0.413 & 0.161 & 0.290 & 0.154 & 0.305 & 0.301 \\
& she & 0.101 & 0.023 & 0.453 & 0.282 & 0.077 & 0.214 & 0.284 & 0.156 & 0.189 & 0.136 & 0.286 & 0.215 \\
& woman & 0.083 & 0.045 & 0.456 & 0.362 & 0.029 & 0.274 & 0.470 & 0.129 & 0.298 & 0.212 & 0.347 & 0.325 \\
& man & 0.098 & -0.028 & 0.374 & 0.337 & 0.029 & 0.251 & 0.584 & 0.072 & 0.362 & 0.237 & 0.388 & 0.339 \\
\multirow{4}{*}{Fasttext Output}
& he & 0.127 & 0.102 & 0.246 & 0.265 & 0.185 & 0.225 & 0.230 & 0.075 & 0.064 & 0.098 & 0.069 & 0.169 \\
& she & 0.187 & 0.184 & 0.267 & 0.241 & 0.258 & 0.251 & 0.328 & 0.204 & 0.216 & 0.190 & 0.107 & 0.239 \\
& woman & 0.325 & 0.320 & 0.369 & 0.392 & 0.292 & 0.485 & 0.444 & 0.302 & 0.236 & 0.255 & 0.243 & 0.446 \\
& man & 0.171 & 0.024 & 0.148 & 0.043 & 0.214 & 0.077 & 0.113 & 0.152 & 0.277 & 0.121 & 0.119 & 0.127 \\
\bottomrule
\end{tabular}
\end{table}

\begin{table}[!b]
\caption{Calculation of associations between Religion and Target words
using word embeddings (Glove, SpaCy, Fasttext).
}
\fontsize{5}{7}\selectfont
\begin{tabular}{llllllllllllllll}\toprule\\

Religion Bias & input & liberal & violent & un educated & dirty &
judgemental & terrorism & terrorist & conservative & violent \\
\multirow{8}{*}{Glove Output} & judaism & 0.445 & 0.224 & 0.145 & -0.100
& -0.027 & 0.246 & 0.180 & 0.451 & 0.224 \\
& christianity & 0.372 & 0.340 & 0.179 & -0.086 & -0.099 & 0.400 & 0.296
& 0.433 & 0.340 \\
& christian & 0.579 & 0.329 & 0.125 & 0.013 & -0.302 & 0.329 & 0.297 &
0.627 & 0.329 \\
& hindutva & 0.295 & 0.216 & 0.219 & 0.174 & 0.136 & 0.152 & 0.098 &
0.294 & 0.216 \\
& islam & 0.394 & 0.470 & 0.189 & 0.165 & -0.127 & 0.623 & 0.528 & 0.482
& 0.470 \\
& jew & 0.458 & 0.303 & 0.406 & 0.208 & 0.017 & 0.242 & 0.279 & 0.440 &
0.303 \\
& muslim & 0.453 & 0.592 & 0.275 & 0.167 & -0.198 & 0.563 & 0.553 &
0.540 & 0.592 \\
& hinduism & 0.231 & 0.248 & 0.235 & -0.124 & -0.025 & 0.165 & 0.098 &
0.249 & 0.248 \\
\multirow{8}{*}{SpaCy Output} & judaism & 0.399 & 0.155 & 0.197 & 0.025
& 0.287 & 0.217 & 0.149 & 0.342 & 0.155 \\
& christianity & 0.562 & 0.345 & 0.404 & 0.026 & 0.442 & 0.387 & 0.309 &
0.513 & 0.345 \\
& christian & 0.487 & 0.222 & 0.406 & -0.007 & 0.334 & 0.251 & 0.254 &
0.461 & 0.222 \\
& hindutva & 0.000 & 0.000 & 0.000 & 0.000 & 0.000 & 0.000 & 0.000 &
0.000 & 0.000 \\
& islam & 0.287 & 0.202 & 0.121 & 0.093 & 0.087 & 0.308 & 0.304 & 0.187
& 0.202 \\
& jew & 0.091 & 0.036 & 0.162 & 0.244 & -0.087 & 0.037 & 0.137 & -0.032
& 0.036 \\
& muslim & 0.357 & 0.297 & 0.305 & 0.180 & 0.107 & 0.339 & 0.369 & 0.294
& 0.297 \\
& hinduism & 0.230 & 0.140 & 0.141 & 0.121 & 0.234 & 0.179 & 0.068 &
0.171 & 0.140 \\
\multirow{8}{*}{Fasttext Output} & judaism & 0.231 & 0.244 & 0.247 &
0.144 & 0.282 & 0.403 & 0.240 & 0.326 & 0.244 \\
& christianity & 0.309 & 0.336 & 0.297 & 0.196 & 0.263 & 0.492 & 0.345 &
0.434 & 0.336 \\
& christian & 0.258 & 0.194 & 0.186 & 0.265 & 0.162 & 0.260 & 0.242 &
0.217 & 0.194 \\
& hindutva & 0.189 & 0.161 & 0.154 & 0.061 & 0.285 & 0.247 & 0.168 &
0.173 & 0.161 \\
& islam & 0.363 & 0.239 & 0.172 & 0.108 & 0.095 & 0.413 & 0.420 & 0.235
& 0.239 \\
& jew & 0.074 & 0.032 & 0.131 & 0.081 & 0.041 & 0.036 & 0.095 & 0.023 &
0.032 \\
& muslim & 0.385 & 0.376 & 0.274 & 0.250 & 0.169 & 0.464 & 0.577 & 0.302
& 0.376 \\
& hinduism & 0.176 & 0.100 & 0.154 & 0.046 & 0.277 & 0.262 & 0.183 &
0.147 & 0.100 \\
\bottomrule
\end{tabular}
\end{table}

\begin{table}[!b]
\caption{Calculation of correlations between Race and Target words using
word embeddings (Glove, SpaCy, Fasttext).
}
\fontsize{5}{7}\selectfont
\begin{tabular}{llllllllllllllll}\toprule\\
Race Bias
 & 
input
 & 
executive
 &
{%

redneck
} &
{%

leader
} &
{%

farmer
} &
{%

engineer
} & 
laborer
 &
{%

teacher
} &
{%

slave
} &
{%

musician
} &
{%

runner
} &
{%

criminal
} & 
homeless
 \\
\midrule
\multirow{7}{*}{Glove Output} & black & 0.289 & 0.183 &
{%
0.402} &
{%
0.435} &
{%
0.207} &
{%
0.053} & 0.376 &
{%
0.403} &
{%
0.358} &
{%
0.349} &
{%
0.373} &
{%
0.325} \\
& caucasian & -0.190 & 0.320 &
{%
0.169} &
{%
0.206} &
{%
-0.007} &
{%
0.098} & 0.154 &
{%
0.243} &
{%
0.159} &
{%
0.082} &
{%
0.034} &
{%
0.106} \\
& asian & 0.312 & -0.012 &
{%
0.373} &
{%
0.276} &
{%
0.143} &
{%
0.012} & 0.249 &
{%
0.239} &
{%
0.155} &
{%
0.369} &
{%
0.277} &
{%
0.203} \\
& african & 0.385 & -0.030 &
{%
0.540} &
{%
0.381} &
{%
0.244} &
{%
0.079} & 0.335 &
{%
0.346} &
{%
0.294} &
{%
0.310} &
{%
0.306} &
{%
0.317} \\
& white & 0.371 & 0.084 &
{%
0.434} &
{%
0.438} &
{%
0.166} &
{%
0.043} & 0.308 &
{%
0.311} &
{%
0.219} &
{%
0.303} &
{%
0.354} &
{%
0.248} \\
& american & 0.472 & 0.099 &
{%
0.522} &
{%
0.487} &
{%
0.539} &
{%
0.181} & 0.529 &
{%
0.522} &
{%
0.530} &
{%
0.417} &
{%
0.468} &
{%
0.282} \\
& european & 0.459 & -0.203 &
{%
0.459} &
{%
0.118} &
{%
0.268} &
{%
-0.076} & 0.209 &
{%
0.230} &
{%
0.181} &
{%
0.394} &
{%
0.318} &
{%
0.114} \\
\multirow{7}{*}{SpaCy Output} & black & 0.057 & 0.339 &
{%
0.138} &
{%
0.202} &
{%
0.022} &
{%
0.244} & 0.068 &
{%
0.282} &
{%
0.090} &
{%
0.093} &
{%
0.256} &
{%
0.310} \\
& caucasian & 0.076 & 0.244 &
{%
0.204} &
{%
0.195} &
{%
0.005} &
{%
0.238} & 0.143 &
{%
0.311} &
{%
0.100} &
{%
0.148} &
{%
0.164} &
{%
0.211} \\
& asian & 0.036 & 0.231 &
{%
0.030} &
{%
0.115} &
{%
0.015} &
{%
0.141} & 0.106 &
{%
0.194} &
{%
0.051} &
{%
-0.009} &
{%
0.062} &
{%
0.146} \\
& african & 0.195 & 0.120 &
{%
0.211} &
{%
0.270} &
{%
0.185} &
{%
0.258} & 0.167 &
{%
0.271} &
{%
0.219} &
{%
0.015} &
{%
0.150} &
{%
0.199} \\
& white & 0.045 & 0.349 &
{%
0.081} &
{%
0.172} &
{%
0.000} &
{%
0.202} & 0.015 &
{%
0.225} &
{%
0.028} &
{%
0.061} &
{%
0.209} &
{%
0.270} \\
& american & 0.247 & 0.157 &
{%
0.239} &
{%
0.227} &
{%
0.199} &
{%
0.256} & 0.192 &
{%
0.248} &
{%
0.201} &
{%
0.053} &
{%
0.231} &
{%
0.238} \\
& european & 0.242 & 0.089 &
{%
0.166} &
{%
0.187} &
{%
0.128} &
{%
0.246} & 0.093 &
{%
0.204} &
{%
0.055} &
{%
0.009} &
{%
0.183} &
{%
0.163} \\
\multirow{7}{*}{Fasttext Output} & black & 0.316 & 0.417 &
{%
0.328} &
{%
0.353} &
{%
0.276} &
{%
0.223} & 0.302 &
{%
0.345} &
{%
0.359} &
{%
0.392} &
{%
0.394} &
{%
0.349} \\
& caucasian & 0.256 & 0.385 &
{%
0.277} &
{%
0.300} &
{%
0.280} &
{%
0.208} & 0.348 &
{%
0.326} &
{%
0.323} &
{%
0.219} &
{%
0.352} &
{%
0.350} \\
& asian & 0.311 & 0.415 &
{%
0.293} &
{%
0.390} &
{%
0.268} &
{%
0.148} & 0.367 &
{%
0.318} &
{%
0.340} &
{%
0.247} &
{%
0.352} &
{%
0.378} \\
& african & 0.278 & 0.349 &
{%
0.298} &
{%
0.380} &
{%
0.149} &
{%
0.261} & 0.314 &
{%
0.345} &
{%
0.351} &
{%
0.264} &
{%
0.279} &
{%
0.490} \\
& white & 0.286 & 0.404 &
{%
0.286} &
{%
0.309} &
{%
0.295} &
{%
0.214} & 0.279 &
{%
0.321} &
{%
0.345} &
{%
0.394} &
{%
0.382} &
{%
0.365} \\
& american & 0.320 & 0.470 &
{%
0.255} &
{%
0.361} &
{%
0.266} &
{%
0.131} & 0.338 &
{%
0.274} &
{%
0.324} &
{%
0.221} &
{%
0.325} &
{%
0.357} \\
& european & 0.308 & 0.276 &
{%
0.273} &
{%
0.262} &
{%
0.189} &
{%
0.135} & 0.306 &
{%
0.231} &
{%
0.251} &
{%
0.258} &
{%
0.323} &
{%
0.327} \\
\bottomrule
\end{tabular}
\end{table}

\end{document}